%% file: main.tex
\documentclass{article}

\usepackage{spconf,amsmath,graphicx,amsfonts,amssymb}

\usepackage{booktabs}
\usepackage{caption}
\usepackage{hyperref}
\usepackage{tablefootnote}
\usepackage{multirow}
\usepackage{diagbox} %
\usepackage[numbers, sort&compress]{natbib}
\usepackage{subfigure}
\usepackage{float}
\usepackage{breqn}
\usepackage{float}
\usepackage{balance}
\usepackage{makecell}
\usepackage{bibspacing}
\usepackage{lipsum}
\usepackage{bm}

\usepackage[hang,flushmargin]{footmisc}  

\newcommand{\loss}[0]{\mathcal{L}}
\title{Mask6D: Masked Pose Priors For 6D Object Pose Estimation}
\name{Yuechen Xie\qquad Haobo Jiang\qquad Jin Xie}
\address{PCA Lab, School of Computer Science and Engineering\\
	Nanjing University of Science and Technology, Nanjing, China
}
\begin{document}
\ninept

\maketitle

\input{texs/0_abstract}
\vspace{-0.3cm}
\input{texs/1_introduction}
\vspace{-0.3cm}
\input{texs/2_related_work}

\vspace{-0.3cm}
\input{texs/3_method}

\vspace{-0.3cm}
\input{texs/4_experiment}

\vspace{-0.3cm}
\input{texs/6_conclusion}

\clearpage
\bibliographystyle{IEEEbib}
\bibliography{strings,refs}
\end{document}

%% file: texs/0_abstract.tex
\begin{abstract}
	Robust 6D object pose estimation in cluttered or occluded conditions using monocular RGB images remains a challenging task. One reason is that current pose estimation networks struggle to extract discriminative, pose-aware features using 2D feature backbones, especially when the available RGB information is limited due to target occlusion in cluttered scenes. To mitigate this, we propose a novel pose estimation-specific pre-training strategy named Mask6D. Our approach incorporates pose-aware 2D-3D correspondence maps and visible mask maps as additional modal information, which is combined with RGB images for the reconstruction-based model pre-training. Essentially, this 2D-3D correspondence maps a transformed 3D object model to 2D pixels, reflecting the pose information of the target in camera coordinate system. Meanwhile, the integrated visible mask map can effectively guide our model to disregard cluttered background information. In addition, an object-focused pre-training loss function is designed to further facilitate our network to remove the background interference. 
Finally, we fine-tune our pre-trained pose prior-aware network via conventional pose training strategy to realize the reliable pose prediction. Extensive experiments verify that our method outperforms previous end-to-end pose estimation methods. 
\end{abstract}
\vspace{-0.1cm}
\begin{keywords}
	Object Pose Estimation, Self-Supervised Learning, Prior Learning
\end{keywords}

%% file: texs/1_introduction.tex
\section{Introduction}
\begin{figure}[t!]
	\centering
	\subfigure[Basic structure of baseline methods\cite{singlestage,gdrn,sopose}]{
		\label{Fig.sub.1}
		\includegraphics[width=0.95\linewidth]{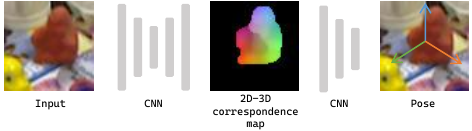}}
		\vspace{-0.2cm}
	\subfigure[Basic structure of our method. ]{
		\label{Fig.sub.2}
		\includegraphics[width=0.95\linewidth]{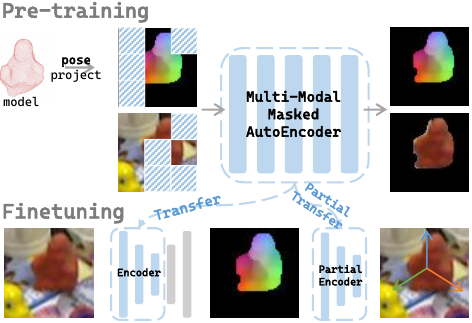}}
	\caption{\textbf{The architecture of end-to-end 6D pose estimation methods.}
		Compared to existing direct regression methods, we propose a novel pre-train method based on MAE\cite{mmae,mae}, which can enable our network to have additional prior knowledge. }
	\label{Fig.teaser}
	\vspace{-0.7cm}
\end{figure}

Estimating the 6D pose of objects (\textit{i.e.} 3D rotation and 3D translation) is a fundamental task in computer vision field, which has extensive applications such as robotic planning and grasping \cite{rotbotm}, autonomous driving \cite{autodrive}, and augmented reality (AR) \cite{ar}. 
Traditional pose estimation methods usually integrate additional depth map into the RGB image to achieve more precise pose estimation. 
In recent years, benefiting from the development of the  deep learning in computer vision field, much more research efforts have been devoted into the learning-based pose estimation methods using pure RGB images. 

The mainstream end-to-end RGB-based pose estimation methods focus on using 2D feature backbone to predict the 2D-3D correspondence map for P\textit{n}P-based pose estimation~\cite{segdriven, pvnet, cdpn}. 
Their key distinction lies in the various forms of the 2D-3D correspondence maps.
For instance, \cite{segdriven, pvnet} use the 2D projections of the target model's key points as their 2D-3D correspondence map. 
Instead, \cite{cdpn, gdrn} propose to leverage the surface coordinates of target model to obtain a dense correspondence representation. 
Also, \cite{sopose} introduces the multi-layer mechanism to establish more robust 2D-3D correspondence map. 
Nevertheless, these 2D backbone-based methods tend to suffer from limited object-pose prior information, which potentially degrades the quality of the predicted 2D-3D correspondences and thus reduce the estimation precision, particularly in some occluded or cluttered settings. 

Inspired by the huge success of Masked AutoEncoder~\cite{mae, mmae}, we propose a novel, pose estimation-specific pre-training framework, named Mask6D, to implicitly enhance the conventional 2D feature backbone with the object-pose prior information. 
Different from traditional Multi-modal MAE~\cite{mmae} that takes depth maps and semantic segmentation maps as input, our Mask6D use the RGB image, 2D-3D correspondence map of the target object and the visible mask map as the input. 
By auto-encoding the masked 2D-3D correspondence maps, we promote the 2D backbone to effectively perceive the potential pose information of the masked object part. 
As such, this pre-trained, pose-aware 2D backbone can be expected to establish more reliable 2D-3D correspondence map in cluttered or occluded settings. 
Furthermore, by taking this visible mask map as the additional input, we can constrain the pre-trained 2D backbone to pay more attention on the promising object region, thereby mitigating the negative interference from the background information in cluttered settings. 
Particularly, in order to further reduce the background interference, we improve the conventional full-pixel pre-training loss function in MAE paradigm with the object-focused, partial-pixel loss function. 

Finally, during the fine-tuning phase,  we follow the traditional end-to-end pose estimation methods to optimize our network.
In detail, we integrate our pre-trained encoder in Mask6D into our pose estimation network for RGB-image feature extraction. In addition, such pre-trained encoder is leveraged as backbone of our regression network for final 6D object pose prediction. 

In summary, our contributions are listed as follows:
\begin{itemize}
	\item We propose a novel, pose estimation-specific pre-training framework, Mask6D, to enhance the 2D feature backbone with the object-pose prior information for more robust pose estimation in cluttered or occluded settings.  
    \item The 2D-3D correspondence map and the visible mask map are innovatively leveraged as the self-supervised signals for promoting our Mask6D to perceive pose prior of the occluded object parts as well as mitigate the background interfence. 
	\item To further compress the background interfence, we enhance the full-pixel pre-training loss function in MAE paradigm with the object-focused, partial-pixel loss function.
\end{itemize}
Extensive experiments conducted on LM~\cite{lm}, LM-O~\cite{lmo}, YCB-V~\cite{posecnn} datasets demonstrate the effectiveness of our proposed method in the 6D object pose estimation task.

%% file: texs/2_related_work.tex
\section{Related Work}
\begin{figure*}[!t]
	\centering
	\includegraphics[width=0.95\textwidth]{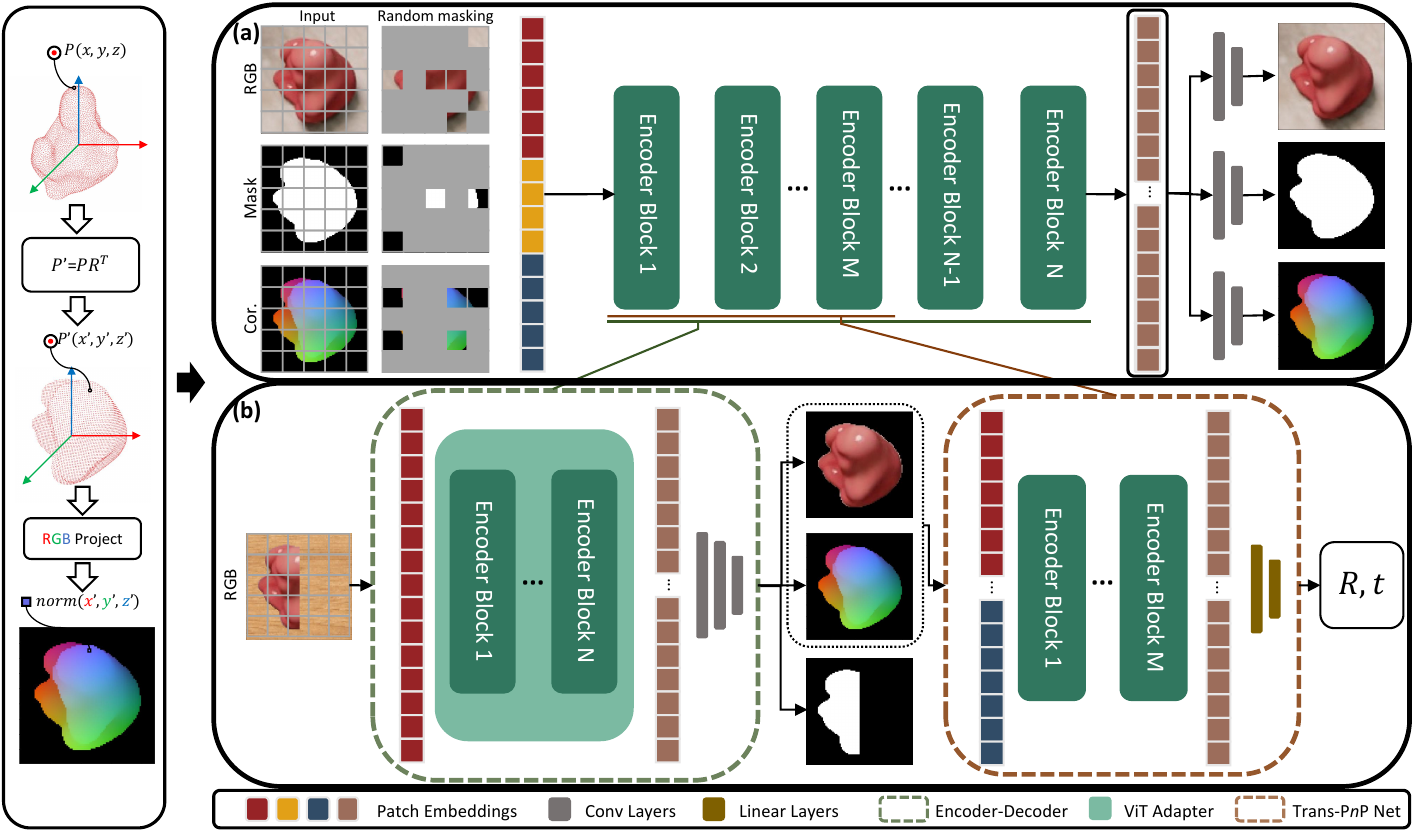}
	\caption{\textbf{Overview of the proposed Mask6D framework. }(a) Pretrain step: Given three modalities of input (i.e., a RGB  image, 2D-3D correspondenses map and visible mask of target object). Then, following the patch-masking strategy of MultiMAE~\cite{mmae}, we randomly select  a subset of  patches from these modalities and learn to reconstruct the  pixel region occupied by the target object. (b) Finetune step: Given an image input of target object, we use the pre-trained encoder with adapter~\cite{vitadapter} as the RGB feature extractor. This allows us to predict the geometric features and complete RGB information of the target. Finally, we utilize the predicted two modalities as inputs and employ some of our pre-trained encoder blocks to directly regresses the 6D object pose. }
	\label{fig:pipeline}
 \vspace{-0.6cm}
\end{figure*}

\textbf{Object pose estimation,} the most widely adopted approach involves a two-stage paradigm, these methods obtain pose parameters by establishing 2D-3D correspondence and then using RANSAC/P\textit{n}P or direct regression. \cite{segdriven,pvnet} aims to establish 2D-3D correspondence through the 2D projection position of key points, and enhance the robustness of the method through voting or other methods, then obtain the pose parameters through RANSAC/P\textit{n}P and its related algorithms. For the second type of methods based on regression, ~\cite{singlestage} innovatively proposed a PointNet-like~\cite{pointnet} pooling network structure that utilizes sparse 2D-3D correspondence to directly regress pose parameters. Additionally, ~\cite{posecnn} decouples the translation parameter into different representations. Inspired by these methods, ~\cite{cdpn} combines direct regression and RANSAC/P\textit{n}P to improve performance. Based on these methods, ~\cite{gdrn} achieved even better performance by improving the parameter representation of pose, which makes the complete direct regression pose parameter method based on dense 2D-3D correspondence more promising. ~\cite{sopose} introduces a multi 3D representation information based on ~\cite{gdrn}. This approach effectively addresses the self-occlusion problem of the target object and achieves significant performance improvement.

\noindent\textbf{Multi-modality self-supervised learning, }it utilizes the inherent features of the data as supervising signals for training. In supervised learning, a large amount of labelled data is usually required to guide model learning, but the cost of data collection and annotation is often high, limiting the applicability of the model. In self-supervised learning, network learns image representation from unlabelled data through designing auxiliary tasks.  These representation prior can be transferred to other tasks, thereby improving the model's performance on these tasks. Inspired by the success of current works related to MAE~\cite{mae}, multi-modal learning~\cite{mmae} achieves more significant performance.

%% file: texs/3_method.tex
\section{Method Description}
Given an RGB image $I$ of the target object captured by an object detector and a set of CAD models of the objects, the 6D object pose estimation task aims to recover the orientation $\mathbf{R}\in SO(3)$ and the spatial position $\mathbf{t}\in \mathbb{R}^3$ of the target object in the camera coordinate system. 
Following the RGB-based pose estimation paradigm, we focus on predicting the 2D-3D correspondence between the 2D object image and the 3D object model for P\emph{n}P-based pose estimation. 
However, due to the severe occlusions and background interference in cluttered environments, robust object pose estimation in cluttered settings is still a challenging task. 

To handle it, inspired by success achieved by the Masked AutoEncoder (MAE), we leverage such masked auto-enconding mechanism for pre-training the 2D backbone in mainstream RGB-based pose estimation methods. The overall pipeline of our Mask6D is presented in Fig.~\ref{fig:pipeline}. 
In pre-training phase, we leverage the 2D-3D correspondence map and the visible object mask as the multi-modal data for masked auto-enconding training. 
Here, the 2D-3D correspondence map is expected to promote the pre-trained 2D backbone to perceive the pose prior information of the masked/occluded object part in occluded settings, while the visible object mask is used to guide the 2D backbone to weaken the negative background inference in cluttered settings (see Sec.\ref{sec:pretrain}). 
In fine-tuning phase, the encoder blocks of the pre-trained 2D backbone are leveraged to construct our RGB feature extractor and the prediction head for final object pose estimation (see Sec.\ref{sec:finetune}).

\vspace{-0.4cm}
\subsection{Learning Masked Pose Priors by Multi-modal Pre-training}\label{sec:pretrain}
\vspace{-0.2cm}
In this section, we will provide a detailed description of our pre-training strategy, as illustrated in Fig.~\ref{fig:pipeline} (a).  

\noindent\textbf{Pose-estimation-specific Multi-modal Input.} 
As demonstrated above, current 2D backbone-based object pose estimation methods still suffer from limited prediction precisions in challenging occluded or cluttered environments. 
The main reason is that these 2D backbones generally suffer from limited pose prior information of occluded object parts, and they are easily interfered by the negative background information in cluttered settings. 
Based on the anaysis above, we propose a pose-aware, multi-modal pre-training framework to mitigate it. 
Specifically, we take the RGB iamge, 2D-3D correspondence map and the visible object map as our pre-training input. 
Then, following the standard masked auto-encoding paradigm, we first tokenize these input data and partially mask them to form a set of mixed multi-modal tokens. 
Then, we perform the encoding and decoding operations on these mixed multi-modal tokens, where the we expect the decoded multi-modal tokens to precisely recover the original RGB iamge, 2D-3D correspondence map and the visible object map. 

The reason for taking the 2D-3D correspondence map and the visible object map as our multi-modal input mainly rely on these two intuitions: (i) to perceive the potential pose prior information
; (ii) to reduce the background interference.

\noindent\textbf{Object-focused Supervision Strategy.}
Although the visible object mask in Mask6D can effectively constrain our pre-trained 2D backbone to pay more attention on the object region, the full-pixel loss function used in conventional MAE potenetially would still unavoidly introduce useless background loss, thereby misleading our model optimziation. 
 To relieve it, we propose an effective, object-focused supervision strategy for more reliable model training. 
Instead of , we just focus on recovering the pixels lying in the object region, which can be formulated as:
\begin{equation}
	\begin{cases}
		\loss_{\text{RGB}}  &= \left\| \bar{M}_{\text{RGB}}\odot \bar{M}_{\text{MASK}}-\hat{M}_{\text{RGB}}\right\|_2, \\
		\loss_{\text{MASK}}  &= \left\| \bar{M}_{\text{MASK}}-\hat{M}_{\text{MASK}}\right\|_1, \\
		\loss_{\text{COR}} &= \left\| \bar{M}_{\text{COR}}\odot \bar{M}_{\text{MASK}}-\hat{M}_{\text{COR}}\right\|_1, 
	\end{cases}
	\label{eq:pretrain_single}
\end{equation}
and 
 \begin{equation}
	\loss_{\text{FOCUS}} = \loss_{\text{COR}}+\loss_{\text{MASK}}+\loss_{\text{RGB}}.
	\label{eq:pretrain_all}
\end{equation}
Here, $\odot$ denotes element-wise multiplication; $\bar{\bullet}$ and $ \hat{\bullet}$ represent the ground-truth map and predicted map, respectively. This loss function achieves object-focused perception by limiting the supervision range from the entire image to the region occupied by the target, thereby enforcing the network to better mine the information of the target itself.

\vspace{-0.3cm}
\subsection{Fine-tuning for Pose Estimation.}\label{sec:finetune}
\vspace{-0.2cm}
Based on the pre-trained 2D backbone, our fine-tuning step then follows the traditional
end-to-end pose estimation methods to fine-tune our network for better adapting specific pose estimation tasks(S).

\noindent\textbf{Robust Multi-modal Prediction}. As demonestrated above, our Mask6D requires multi-modal data as the input for masked auto-encoding training. 
However, during our fine-tuning phase, we can just access to RGB image. We find that such significant input gap between the pre-training and fine-tuning phases would largely weaken the effectiveness and performance of our framework. 
To mitigate this input gap, given a RGB image, we advocate for using the pre-trained encoder blocks from our Mask6D to first predict the associated 2D-3D correspondence map and the visible object map. Owing to our effective pre-training strategy, we can  achieve high-quality predictions of these correspondence map and visible object map. 
Then, we feed the RGB image and the predicted correspondence map into a pre-trained encoder-based pose regression network for final object pose estimation. 
The formal loss function can be written as below:
\vspace{-0.2cm}
\begin{equation}
	\loss_{\text{ours}} = 1-SSIM\footnote{SSIM: Structural Similarity Index Measure\cite{ssim}.}(\bar{M}_{\text{RGB}}\odot\bar{M}_{\text{MASK}},\hat{M}_{\text{RGB}}\odot\bar{M}_{\text{MASK}} )+\loss_{\text{GDR}},
	\label{eq:pretrain_single}
\end{equation}
where $\loss_{\text{GDR}}$ denotes a combined loss term of correspondence map $\bar{M}_{\text{COR}}$, visible mask $\bar{M}_{\text{MASK}}$, and surface region attention $\bar{M}_{\text{SRA}}$ from GDR-Net~\cite{gdrn}.

\begin{figure}[t]
	\centering
	\includegraphics[width=0.95\columnwidth]{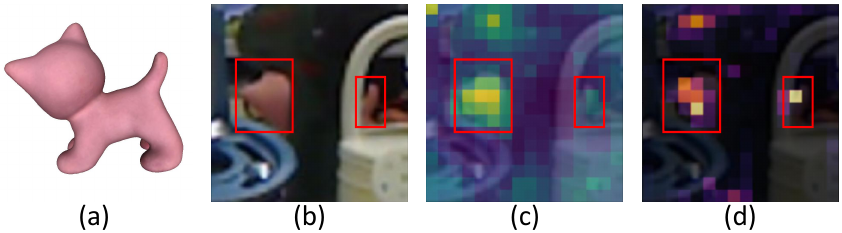}
	\caption{\textbf{Comparison of pre-trained ViT attention map and CNN feature map, }(a) target object model, (b) truncated target image,(c) $16\times16$ feature map from baseline CNN-based encoder, (d) attention map from our pre-trained encoder.}
	\label{fig:atten_vis}
 \vspace{-0.5cm}
\end{figure}

To demonstrate the effectiveness of our method, we visualized the feature map of baseline's encoder~\cite{gdrn} and attention map of our pre-train encoder in cluttered scenarios in Fig.\ref{fig:atten_vis}. It is evident that, thanks to the attention mechanism and pre-training, our encoder can capture the features of truncated target more efficiently, thereby enhancing the performance of our method in difficult scenarios.

\noindent\textbf{Pose Parameter Regression. }In the previous pre-training steps, transformer blocks in our pre-trained  encoder had  obtained various types of prior needed for pose estimation, therefore, we are considering how to make efficient use of the existing prior knowledge.

During pre-training, our encoder takes the target's RGB, visible mask, 2D-3D correspondence as input. It encompasses the input of the P\textit{n}P network of the existed method~\cite{gdrn}. Therefore, we directly replace the tradition random initialized convolutional network of P\textit{n}P network with the first 'M' blocks of the encoder network. Considering the balance between computational cost and performance, we do not separately process the decoder's output into different class tokens for each modality. Instead, we concatenate them directly and fuse them through a convolutional layer, and process the fused feature map into tokens as input of our Trans-P\textit{n}P, our method is shown in Fig.~\ref{fig:pipeline}.b.
\begin{equation}
\mathbf{P}=\mathrm{Trans}\mathrm{-P}n\mathrm{P}(\mathrm{Conv(}{\mathrm{cat}(M_{\text{SRA}}, M_{\text{COR}}, M_{\text{RGB}}))})
\end{equation}
The supervision signals in this section refer to the definition in GDR-Net~\cite{gdrn}, use  a loss term combined with translation parameters and point cloud matching.

%% file: texs/4_experiment.tex
\section{Experiments}
\vspace{-0.4cm}
\subsection{Experiments Setup}
\vspace{-0.2cm}
\noindent\textbf{Implementation Details.} 
All our experiments are implemented by using PyTorch~\cite{pytorch}. 
We use a batchsize of 48 and about 400k steps on each datasets for pre-training. The other settings during pre-training and fine-tuning are the same as GDR-Net~\cite{gdrn}.
The number of blocks ('M' in Fig.~\ref{fig:pipeline}) in Trans-P\textit{n}P is set to 3, 4, and 5 for LM, LM-O,YCB-V depending on the difficulty of the datasets. Meanwhile, ViT-adapter \cite{vitadapter} is introduced to solve the problem of inductive bias missing in dense prediction tasks of ViT\cite{vit} based encoder. Input resolutions are set to 256*256, 64*64, 64*64 for RGB, mask, and coordinate respectively by following baseline method, and patch sizes are set to 16*16, 4*4, 4*4 to ensure each modality comprises 256 patches. 

\noindent\textbf{Datasets.} We conduct out method on three commonly-used datasets, including LineMod\cite{lm}(LM), LineMod-Occluded\cite{lmo}(LM-O),  YCB-Video\cite{posecnn}(YCB-V). Among these, LM-O comprises a LM sequence with more occlusion, while YCB-V is a challenging dataset with significant occlusion and clutter.
On the pre-training phase, we use the same data configuration as the fine-tuning stage, and the specific configuration is the same as \cite{gdrn,sopose}. Considering efficiency and simplicity, we pre-train on LM, LM-O together, and pre-train on YCB-V separately.

\noindent\textbf{Evaluation Metrics.} We use common metrics ADD(-S)~\cite{metrics1,metrics2}. The ADD metric~\cite{metrics1} measures whether the average deviation of the transformed model points is less than 10\% of the object's diameter (0.1d).

\vspace{-0.2cm}
\subsection{Ablation study.}
\vspace{-0.2cm}
\input{tables/lm_ablation}

In Table.~\ref{tb:lm_result}, we demonstrate our results for LM datasets with respect to the ADD(-S) metric. We can see that our method has significant improvements over other methods on various metrics.
At the same time, our ablation study show the effectiveness of our proposed method.
\vspace{-0.2cm}
\subsection{Comparison.}
\vspace{-0.2cm}
\input{tables/lmo}

\input{tables/ycb}

In this section, we selected some end-to-end, correspondence based, iterative refinement methods for comparison. We compare our approach with others on the LM-O~\cite{lmo} and YCB-V~\cite{posecnn} datasets.

\noindent\textbf{Results on LM-O. }We compare our method with other methods in terms of ADD-(S) in Table.~\ref{tb:lmo_result}.
Our method leads the use of two-layer representation method SO-Pose (\textbf{65.2} against 62.3) even when using the same single-layer representation as GDR-Net, and significantly outperforms other methods.

\noindent\textbf{Results on YCB-V.} For YCB-V, we show our results on Table.~\ref{tb:ycb}, we outperform again all other methods under ADD-(S) and AUC of ADD-(S) metrics. It is slightly inferior to CosyPose~\cite{cosypose} since it is an iterative refinement method, and our method only requires single forward pass to obtain the pose.

%% file: tables/lm_ablation.tex
\begin{table}[h]
	\footnotesize
	\centering
	\resizebox{1\columnwidth}{!}{
		\centering
		\begin{tabular}{c|l|c c c}
			\toprule
			 \multirow{2}*{Row} &\multirow{2}*{Method} & \multicolumn{3}{c}{ADD(-S)}
			 \cr  &  &0.02d &0.05d &0.10d 
			\tabularnewline
			\hline
		
			A0 &CDPN\cite{cdpn}     & -     & -    & 89.9        \tabularnewline
			A1 &GDR-Net\cite{gdrn}  & 35.3     & 76.3    & 93.7       \tabularnewline
   			A2 &SO-Pose\cite{sopose}  & 45.9     & 83.1   & 96.0        \tabularnewline
   \hline
						B0 &Mask6D (\textbf{Ours})  &\textbf{48.1} &\textbf{86.0} &\textbf{97.6}      \tabularnewline

      \hline
			C0 &B0:\space $\mathcal{L}_{\mathrm{FOCUS}} \rightarrow w/o    \  \mathcal{L}_{\mathrm{MASK}}$  &44.3 &83.4 &96.7        \tabularnewline
   			C2 &B0:\space $\mathcal{L}_{\mathrm{FOCUS}} \rightarrow \mathcal{L}_{\mathrm{MAE}}$   &42.2 &82.0 &96.1   \tabularnewline
   			\hline
      		D0 &B0:\space   $\mathcal{L}_{\mathrm{ours}} \rightarrow \mathcal{L}_{\mathrm{GDR}}$   &42.4         &83.2 &96.5  \tabularnewline
   			D1 &B0:\space Trans-P\textit{n}P $\rightarrow$ Patch-P\textit{n}P\cite{gdrn}   &47.6   &85.5&97.4 \tabularnewline
	\bottomrule
	\end{tabular}
}
\caption{~\label{tb:lm_result}\textbf{Ablation Study on LM. }We provide results of our method with pre-training settings. Block C modifies the strategy for pre-training phase, block D for the fine-tuning phase.
}
\end{table}

%% file: tables/lmo.tex
\vspace{-0.3cm}
\begin{table}[h!]
	\footnotesize
	\centering
	\resizebox{1\columnwidth}{!}{
		\centering
		\begin{tabular}{c|c|c|c}
			\toprule
			 Method &Training Data&P.E. &ADD(-S)
    \tabularnewline
    \hline

    PoseCNN\cite{posecnn}         &\emph{real+syn}   &1 &24.9
    \tabularnewline
    PVNet\cite{pvnet}           &\emph{real+syn}   &N &40.8
    \tabularnewline
    Single-stage\cite{singlestage}    &\emph{real+syn}   &N &43.3
    \tabularnewline
    HybridPose\cite{hybridpose}      &\emph{real+syn}   &N &47.5
    \tabularnewline
    \hline
    GDR-Net\cite{gdrn}         &\emph{real+syn}   &1 &47.4
    \tabularnewline
    GDR-Net\cite{gdrn}         &\emph{real+pbr}   &1 &56.1
    \tabularnewline
    SO-Pose\cite{sopose}         &\emph{real+pbr}   &1 &62.3
    \tabularnewline
    \hline
    Ours &\emph{real+pbr}   &1 &\textbf{65.2}

    \tabularnewline
    \bottomrule
	\end{tabular}
 }
 \caption{~\label{tb:lmo_result}\textbf{Comparsion on LM-O. }We report the Average Recall(\%) of ADD(-S). \emph{pbr} means physically-based rendering. P.E. means whether the method is trained with 1 pose estimator for the whole dataset or 1 per object (N object in total). We use \textbf{bold} to indicate the best result. 
}
\vspace{-0.5cm}
\end{table}

%% file: tables/ycb.tex
\vspace{-0.3cm}
\begin{table}[!htbp]
	\footnotesize
	\centering
	\resizebox{1\columnwidth}{!}{
		\centering
		\begin{tabular}{c|c|c|c|c|c}
			\toprule
			 \multirow{2}*{Method} &\multirow{2}*{P.E.} &\multirow{2}*{Ref.} &ADD &AUC of &AUC of
    \tabularnewline
    &&&(-S)&ADD-S &ADD(-S)
    \tabularnewline
        \hline
    PoseCNN\cite{posecnn}            &1 &&21.3&75.9&61.3
    \tabularnewline
    SegDriven\cite{segdriven}            &1& &39.0&-&-
    \tabularnewline
    PVNet\cite{pvnet}            &N&&-&-&73.4
    \tabularnewline
    S.Stage\cite{singlestage}       &N& &53.9&-&-
    \tabularnewline
    DeepIM\cite{deepim} &1 &\checkmark &-&88.1&81.9
    \tabularnewline
    CosyPose\cite{cosypose} &1 &\checkmark &-&89.8&\textbf{84.5}
    \tabularnewline
    \hline
    GDR-Net\cite{gdrn}            &1& &49.1&89.1&80.2
    \tabularnewline
    SO-Pose\cite{sopose}            &1& &56.8&90.9&83.9
    \tabularnewline
    \hline
    Ours   &1& &\textbf{59.5} &\textbf{91.5} &83.5
    \tabularnewline
    \bottomrule
	\end{tabular}
 }
 \caption{~\label{tb:ycb}\textbf{Comparsion on YCB-V.} Ref. stands for refinement. P.E. reflects the training strategy of pose estimator, 1 represents single model for all objects while N represents one model per object.
}
\vspace{-0.3cm}
\end{table}

%% file: texs/6_conclusion.tex
\section{Conclusion}
\vspace{-0.3cm}

In this study, we introduce an innovative pre-training approach for the object 6D pose estimation task. Our method utilizes 2D-3D correspondences, visible masks, and RGB images as multi-modal inputs for self-supervised pre-training. This approach aids our network in acquiring task-specific prior information for subsequent object pose estimation tasks. Finally, building upon the pre-training, we establish an end-to-end 6D pose regression framework, which demonstrates a substantial improvement compared to other end-to-end methods.